\documentclass[citeauthoryear]{asai}
\usepackage[T1]{fontenc}
\usepackage[latin1]{inputenc}
\usepackage{a4}
\usepackage{graphics}
\usepackage{subfigure}
\IfFileExists{url.sty}{\usepackage{url}}
                      {\newcommand{\url}{\texttt}}

\makeatletter

\providecommand{\LyX}{L\kern-.1667em\lower.25em\hbox{Y}\kern-.125emX\@}
\newenvironment{LyXParagraphIndent}[1]%
{
  \begin{list}{}{%
    \setlength\topsep{0pt}%
    \addtolength{\leftmargin}{#1}
    \setlength\parsep{0pt plus 1pt}%
  }
  \item[]
}
{\end{list}}
\let\SF@@footnote\footnote
\def\footnote{\ifx\protect\@typeset@protect
    \expandafter\SF@@footnote
  \else
    \expandafter\SF@gobble@opt
  \fi
}
\expandafter\def\csname SF@gobble@opt \endcsname{\@ifnextchar[%]
  \SF@gobble@twobracket
  \@gobble
}
\edef\SF@gobble@opt{\noexpand\protect
  \expandafter\noexpand\csname SF@gobble@opt \endcsname}
\def\SF@gobble@twobracket[#1]#2{}

 \newenvironment{lyxcode}
   {\begin{list}{}{
     \setlength{\rightmargin}{\leftmargin}
     \raggedright
     \setlength{\itemsep}{0pt}
     \setlength{\parsep}{0pt}
     \verbatim@font}%
    \item[]}
   {\end{list}}

\usepackage{chicago}
\usepackage{openbib}
\usepackage{times}
\usepackage{mathptm}
\usepackage{dcolumn}

\makeatother

\begin{document}

\title{Brainstorm/J: a Java Framework\\
for Intelligent Agents}

\author{A.~Zunino and A.~Amandi}

\institute{ISISTAN Research Institute}

\institute{Fac. de Ciencias Exactas, Dpto. Computación y Sistemas\\
Universidad Nacional del Centro\\
Campus Univ. Paraje Arroyo Seco - (7000) Tandil - Bs. As., Argentina \\
email: \texttt{\{azunino, amandi\}@exa.unicen.edu.ar}~\\
www: \url{http://www.exa.unicen.edu.ar/~isistan}}

\maketitle
\begin{abstract}
Despite the effort of many researchers in the area of multi-agent systems (MAS)
for designing and programming agents, a few years ago the research community
began to take into account that common features among different MAS exists.
Based on these common features, several tools have tackled the problem of agent
development on specific application domains or specific types of agents. As
a consequence, their scope is restricted to a subset of the huge application
domain of~MAS. In this paper we propose a generic infrastructure for programming
agents whose name is Brainstorm/J. The infrastructure has been implemented as
an object oriented framework. As a consequence, our approach supports a broader
scope of MAS applications than previous efforts, being flexible and reusable.
\end{abstract}
\begin{LyXParagraphIndent}{1em}

\end{LyXParagraphIndent}

\section{Introduction}

Intelligent agents are one of the most rapidly developing areas of research
of the last years. Agents offers new ways to analyze, design and implement software
systems, improving, potentially, the ways in which software is modeled and materialized.

Applications of intelligent agents are surprising varied: from simple personal
assistants~\cite{mitchell94:personAssis}, mobile robots~\cite{brooks86:subsumption}
and Internet agents~\cite{etzioni94:softbot}, to complex autonomous propulsion
systems for the space shuttle~\cite{georgeff89:prs}, assistants for software
development~\cite{ortigosa99:smartBooks} and agents for process scheduling
in industrial systems~\cite{fischer94:reactSched}. On the other hand, agents
can carry out this variety of activities in different ways~\cite{wooldridge98:agentSE}:
interaction with people and/or agents, reactivity, deliberation, representation
and manipulation of mental states, and mobility, among others. 

The construction of agents involves several steps from their conception to the
execution of code. In order to make easier that process, models~\cite{rao91:modeling,shoham93:aop},
architectures~\cite{Muller96:DIA,georgeff89:prs,amandi98:metaArchitecture}
and methodologies~\cite{demazeau98:methodology,drogoul98:methodology,wooldridge99:methodology}
have been proposed. 

Models define a general structure and the relationships among the main components;
software architectures gives more detail, specifying how the components interact
through specific protocols; finally, methodologies define steps to follow using
models, architectures and so on for building applications.

Concerning to architectures, many agent architectures have been proposed in
the literature. Several of them can be classified as software architectures
since they prescribe an implementation. We can mention Interrap~\cite{Muller96:DIA,fischer96:pragmatic_bdi},
ARCHON~\cite{jennings94:archon,cockburn96:archon}, and Brainstorm~\cite{amandi97:ooap,amandi98:metaArchitecture}
among them.

Software architectures~\cite{shaw96:architecture} prescribe an implementation
as a specification of components and interfaces. Developers using these software
architectures implement the components and then connect them through the specified
interfaces. Thus, developers have a reusable design in a high level of abstraction
to develop their multi-agent systems. 

Taking an architecture as a base, a code scheme can be built providing to the
developer the common code of those architectural components. This code scheme
is named \emph{framework}~\cite{johnson91:reusinOOD} by the object-orientation
community.

A framework is defined as a set of classes that implement the common flow of
control among objects in a schematic way. A framework of agents allows the construction
of multi-agent systems by subclassification and composition of this set of classes.
Classes built by subclassification implement the behavior related to a particular
application being built. Moreover, the functionality provided by a framework
can be extended by means of subclassification.

In this paper, we present a framework for multi-agent systems named Brainstorm/J
that is based on the Brainstorm architecture~\cite{amandi97:ooap,amandi98:metaArchitecture}.
This architecture allows simple objects to be extended to become agents. The
extension is supported by meta-objects~\cite{maes87:reflection}, which represent
a flexible and adaptable way for composing components.

The paper is organized as follows. Section~\ref{sec:Brainstorm} briefly describes
the Brainstorm architecture. Section~\ref{sec:Brainstorm/J} presents the framework
Brainstorm/J. In this presentation we show how a new multi-agent application
reuses both common components and control flow specified in the framework. Section~\ref{sec:Experiences}
discuss our experiences with the framework. After that, Sect.~\ref{sec:Related}
describes some related work. Finally, Sect.~\ref{sec:Conclusions} presents
the conclusions.

\section{Brainstorm\label{sec:Brainstorm}}

The backbone of Brainstorm/J is a class hierarchy with a small number of abstract
classes that have been built based on the design of the Brainstorm architecture~\cite{amandi97:ooap,amandi98:metaArchitecture}.
This section briefly describes the organization of the architecture and its
components.

The conception of the Brainstorm architecture is supported on the fact that
a multi-agent system~(MAS) can be considered as an object-oriented system that
has associated a meta-system~\cite{amandi97:ooap,amandi98:metaArchitecture}.
This meta-system incorporates typical agent behavior to simple objects. Then,
an agent is considered as an object with a layer of intelligence in the meta-level.
Thus capabilities such as communication, perception, reaction and deliberation,
that are not inherent to objects, can be introduced in a meta-level.

For providing agent capabilities to objects, the Brainstorm architecture prescribes
the usage of meta-objects. A meta-object~\cite{maes87:reflection} is a special
object that has the ability of intercepting the invocation to methods, altering
thus their execution. In this way, a multi-agent system is defined as an object-oriented
system in which some objects that are considered agents have associated a meta-level
with agent capabilities. The set of meta-levels of the objects considered agents
compose a meta-system. This meta-system is a reflective system causally connected
to the object-oriented system defined in the base-level. The causal connection
is established when an object receives a message. If the object has been defined
as an agent, the message is intercepted by the meta-level, which decides what
to do with the message. In this way, for example, an agent can decide to deny
a request of another agent or to initiate a planning algorithm to achieve its
goals.

Brainstorm allows each object to have several meta-objects associated. The amount
of meta-objects associated to an object depends on the required agent functionality.
Therefore, the selection of a set of meta-objects for an object allows the developer
to define different types of agents such as reactive, deliberative or hybrid
agents.

\begin{figure}
{\par\centering \resizebox*{0.95\columnwidth}{!}{\rotatebox{-90}{\includegraphics{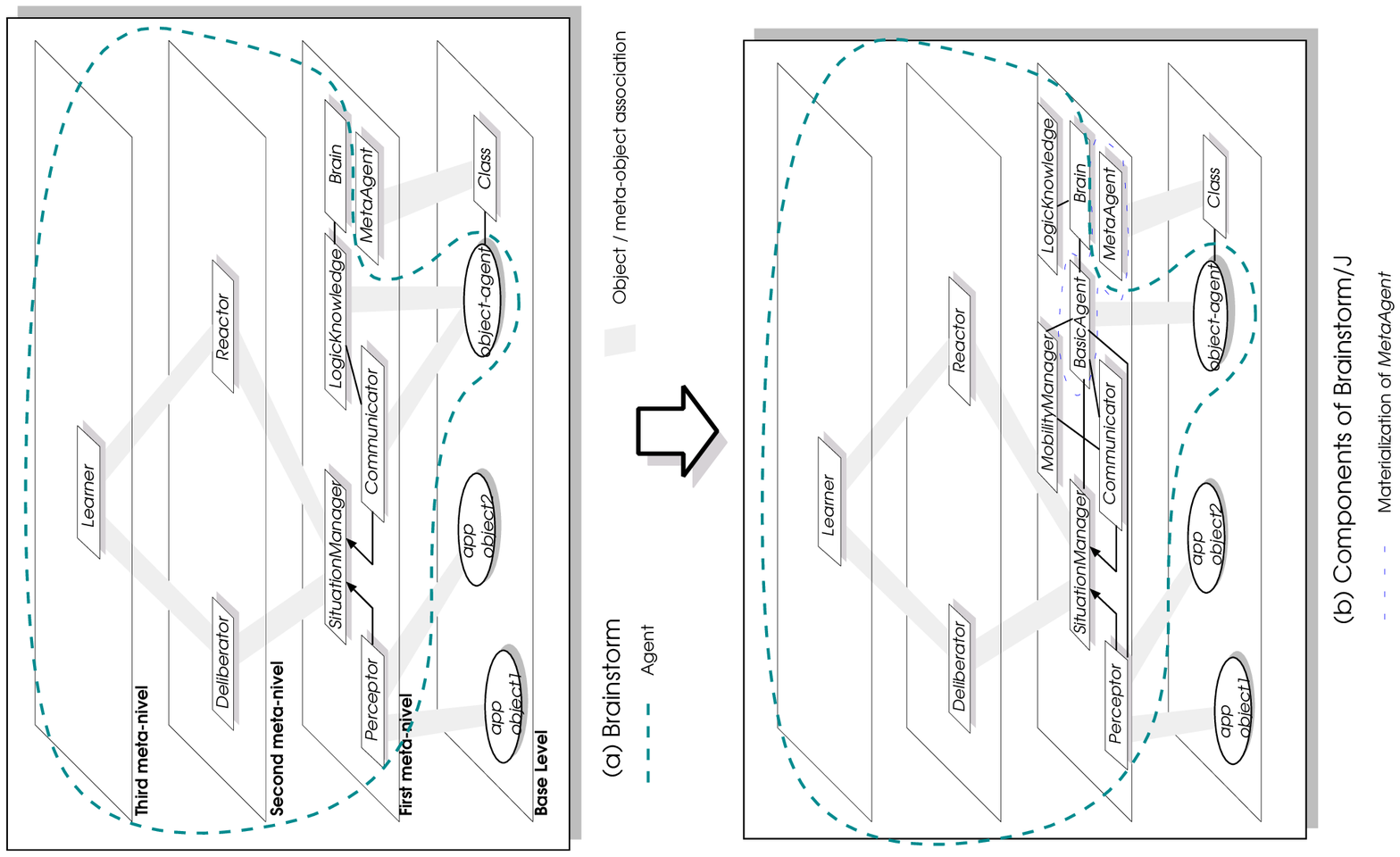}}} \par}

\caption{Architectural materialization\label{fig:materialization}}
\end{figure}

Figure~\ref{fig:materialization}(a) shows a scheme of the architecture. We
can observe that an object has several meta-objects in different levels. In
the first level, four types of meta-objects are defined: creation, perception,
communication and knowledge meta-objects. In the second level, two types of
meta-objects reflect the behavior of the meta-objects of the first level: reaction
and deliberator meta-objects. Finally, in the last level, learner meta-objects
can be added.

The communication system (messages) is uniform enabling objects and agents to
work together. Interaction among agents is established through messages or indirectly
through perception. An object acquires these capabilities by associating meta-objects
to it.

The communication meta-object defines the communication language used by an
agent. The messages received by the agent's object are intercepted by the associated
communication meta-object. In this way, any object can acquire the capability
of using an agent communication language such as KQML or FIPA's ACL.

An agent can also perceive changes in its environment. This is achieved by means
of a set of perception meta-objects. A perception meta-object observes the behavior
of an agent or an object, detecting the invocation to methods. Thus, it records
any event of interest. For example, an agent~\( A \) would like to perceive
the communications received by an agent~\( B \) from agent~\( C \). Then,
agent~\( A \) defines a perception meta-object for~\( B \), indicating what
it shall observe. Thus, an agent can transparently perceive events taking place
in its environment.

The knowledge meta-object is responsible for mental states. It provides facilities
for managing knowledge expressed as logic clauses. This capability is possible
since an integration of an object-oriented language with a logic language has
been defined~\cite{amandi99:multiParadigm}. The goal of combining objects
and logic is to provide a support for allowing objects to define and use mental
states represented as logic clauses. Moreover, mental states can be represented
as objects, clauses or clauses composed by objects. The combination allows objects
to: define mental states in instance variables, define mental states in methods,
refer to mental states in methods, represent mental states as logic clauses
and objects, and inherit mental states among classes.

Agents define their behavior by associating meta-objects of type reactor and/or
deliberator. Reactors and deliberators intervene in all messages that support
interaction. When an agent perceives something (through some perception meta-object)
or receives a message (intercepted by its communication meta-object), an interesting
situation can be detected. Then, the reactor and/or the deliberator meta-objects
intervene, reacting and/or deliberating.

The reaction of an agent involves the immediate execution of different activities
that can be basic actions defined in the associated object, a change in agents'
beliefs, a message to other agent, etc. An agent's deliberative process involves
carefully thinking before making a decision about what to do next.

In the last level of the architecture, a learner component is defined. Meta-objects
intervene in deliberative processes to help in decisions. Regarding to the reaction
process, these meta-objects can alter some reactions.

\section{The Framework Brainstorm/J\label{sec:Brainstorm/J}}

For developing multi-agent systems, it is necessary to have an appropriate agent-oriented
support~\cite{jennings98:aa-mas}. Additionally, it is very convenient to make
use of software components that permit to compose complex agent behaviors~\cite{amandi97:ooap}
to build different types of agents. To achieve these goals we have developed
an object-oriented framework based on the Brainstorm agent architecture. 

The Brainstorm architecture prescribes mechanisms for defining agents through
a set of meta-objects. Based on Brainstorm we have defined a set of agent components
by means of Java classes that can be specialized for building particular agents.

The framework has been implemented by using JavaLog, a multi-paradigm language
based on Java and Prolog~\cite{amandi99:multiParadigm}, and a meta-object
support for Java based on LuthierMOPS~\cite{campo99:luthier}.

This section is structured as following. First, we describe the materialization
of the architecture in the framework. Then, we define details of several classes
of the framework.

\subsection{Mapping architectural components into classes}

The framework Brainstorm/J~\cite{zunino2000:brainstormJ} has been built from
the Brainstorm architecture. Figure~\ref{fig:materialization} shows the materialization
of the architecture. It presents the mapping from architectural components to
classes. 

Figure~\ref{fig:materialization}(a) shows a basic scheme of the architecture.
In Fig.~\ref{fig:materialization}(b) several direct mapping from architectural
components to classes are visible: a base object, a situation manager, and several
meta-objects such as perceptors, deliberators, reactors, learners and meta-agents. 

Architectural components related to communication and knowledge management are
materialized by using more than one class. The communication component is materialized
by two classes, a communicator meta-object and a mobility object. 

The knowledge management represented by the architectural components \emph{LogicKnowledge}
and \emph{Brain} is materialized by means of the multi-paradigm language JavaLog~\cite{amandi99:multiParadigm}.
This language integrates object-oriented programming with logic programming.

\subsection{Creating an agent from an object}

Brainstorm prescribes an architectural component named \emph{MetaAgent}, which
is responsible for agent creation and initialization. This component is able
to create different types of agents (reactive, deliberative, hybrid, etc.) with
diverse agent capabilities (reaction, perception, communication, etc.).

In Brainstorm/J, \emph{MetaAgent} has been materialized by two classes: \texttt{MetaAgent}
and \texttt{BasicAgent}. The first one is in charge of agent creation. The second
one is responsible for initializing the components of an agent components and
keeping information about it.

Agents built with Brainstorm/J are composed by an object living on the base-level,
and some objects and meta-objects situated on meta-levels. The object situated
on the base-level represents an agent's basic skills or capabilities. These
skills can be considered as the effectors of the agent or its devices to modify
the environment. Note that these skills do not include intelligence at all,
since intelligence is added by associating meta-objects. In order to create
agents, every \texttt{MetaAgent} meta-object is associated to some class situated
on the base level. Then, objects belonging to classes with an associated \texttt{MetaAgent}
are \emph{agentified} immediately before their creation.

\begin{figure}
{\par\centering \subfigure[The FORKS application\label{fig:forklifts}]{\resizebox*{0.3\columnwidth}{!}{\includegraphics{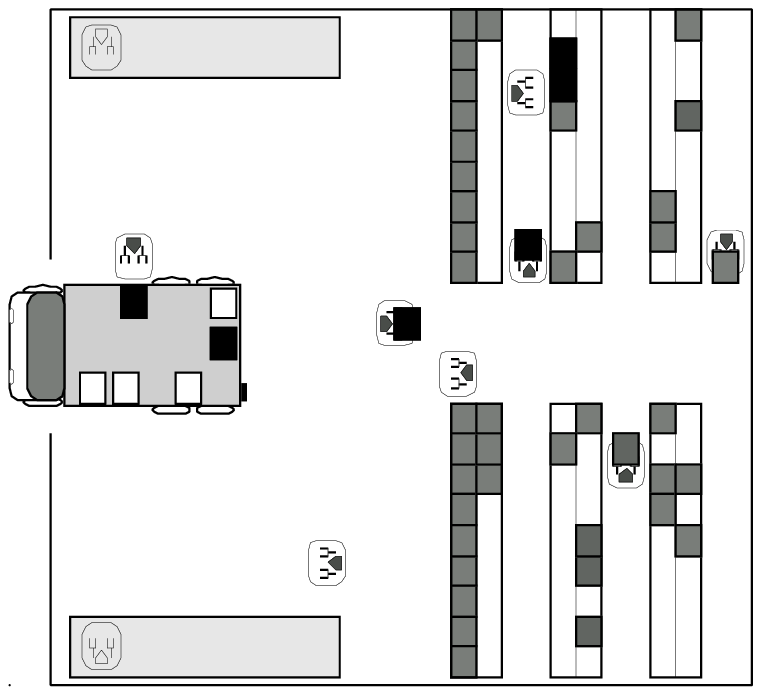}}} ~~~~~\subfigure[Agent creation\label{fig:agentCreation}]{\resizebox*{0.6\columnwidth}{!}{\includegraphics{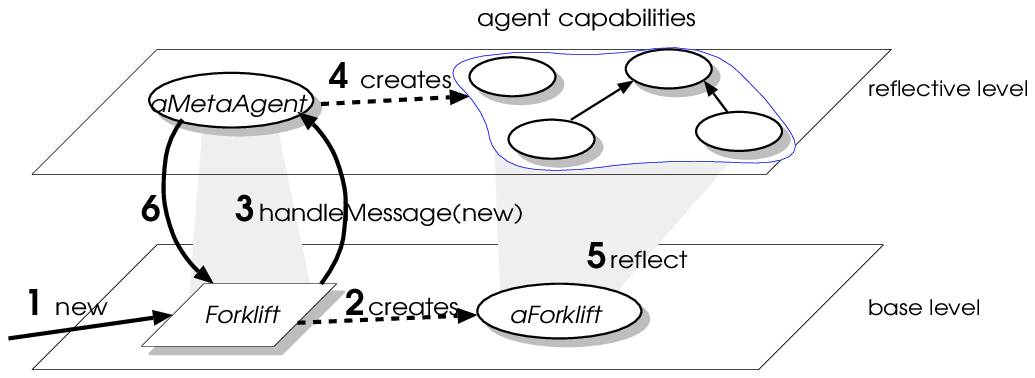}}} \par}

\caption{The FORKS multi-agent system}
\end{figure}

To clarify the process of agent creation we will describe a multi-agent system
named FORKS~\cite{Muller96:DIA} which has been implemented with Brainstorm/J.
The system consists of a set of forklift robots, which try to move a number
of boxes from a truck to some shelves, as shown in Fig.~\ref{fig:forklifts}.
Each robot has a number of basic skills such as advance, turn to some direction,
grasp a box situated in its front, put a box and perceive its environment. These
skills are represented as methods of class \texttt{Forklift}. It is worth noting
that this class represents only the basic skills of a forklift. That is, objects
of this class are not agents, since they do not have agent capabilities.

In order to create agents having instances of \texttt{Forklift} as base-objects,
we have associated a meta-object of class \texttt{MetaAgent} to \texttt{Forklift}.
Figure~\ref{fig:agentCreation} shows a meta-object called \texttt{aMetaAgent}
associated to the class \texttt{Forklift}. This meta-object is activated when
\texttt{Forklift} receives a message for creating a new instance, for example
\texttt{aForklift}. The activation of the meta-object triggers the creation
of a set of objects and meta-objects defining agent capabilities such as perception,
reaction, communication, mobility, deliberation, etc., depending on the desired
agent capabilities. Then, these objects and meta-objects are initialized by
an instance of \texttt{BasicAgent}. Finally, the meta-objects are associated
to the base-object \texttt{aForklift}, adding agent capabilities to it.

Since objects and meta-objects which belong to the reflective level are in charge
of agent capabilities, some of them can be optional. For example, a reactive
agent does not use a deliberator object since its reactive behavior is encoded
into a reactor object.

In order to support different types of agents, \texttt{MetaAgent} provides a
number of methods which can be used by the programmer to specify the capabilities
of agents to be created. For example, the \texttt{hasReaction} method invoked
with a \texttt{true} argument means that agents created by a \texttt{MetaAgent}
meta-object have reactive capability\footnote{%
Instances of \texttt{MetaAgent} can be considered as a class of agents.
}.

The basic idea to capture the agent creation process in an application independent
way consists of implementing that functionality into a so called \emph{template}
method. A template method~\cite{johnson91:reusinOOD,fayad97:frameCACM} is
a method that specifies an operation in an application independent way, describing
the flow of control among a set of objects. The feature that distinguishes a
template method from other types of methods is the fact that a template method
always invokes to several \emph{abstract} methods.

An abstract method can be considered as a \emph{hole} into the framework, which
should be completed by the programmer with the application dependent functionality.
It is worth noting that template methods provides a way to program operations
in an application independent way. At the same time, they can be customized
by defining abstract methods.

As an example of a template method, Fig.~\ref{fig:agentCreation} shows a class
diagram with a template method named \texttt{createAgent}. It invokes the abstract
method \texttt{initCommunication} declared in the abstract class \texttt{Communicator}.
In a concrete subclass of \texttt{Communicator} named \texttt{KQMLCommunicator},
the \texttt{initCommunication} method is defined.

\begin{figure}
{\par\centering \includegraphics{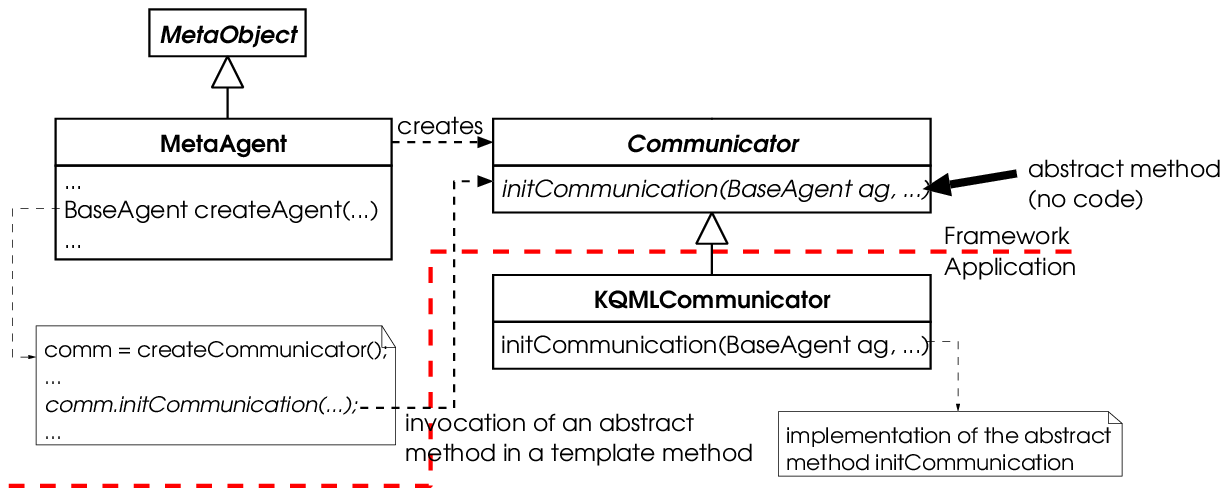} \par}

\caption{The \texttt{createAgent} template method\label{fig:createAgent}}
\end{figure}

\subsection{Perception}

Perception capabilities of agents are implemented by class \texttt{Perceptor}.
This class is a type of meta-object, thus its instances are meta-objects capable
of intercepting messages between objects transparently. Note that perception
usually refers to detecting changes in the environment. Here, the environment
as well as other agents are represented by objects and meta-objects. Therefore,
it is possible to detect changes in these entities by intercepting message between
objects.

The class \texttt{Perceptor} defines the common functionality of perception.
A meta-object of this class is notified with \texttt{messagePerceived} when
an interesting message is received by one of its perceived objects. Programmers
can redefine that method to process each perceived message according to their
requirements. By default, that method notifies to a \texttt{Situation Manager}
object about every perception, giving it opportunity to detect interesting situations.
A typical usage of perception consists of redefining \texttt{messagePerceived}
to add and update agents' beliefs according to perceived messages. 

While other approaches for perception based on the \emph{observer} design pattern~\cite{gamma95:patterns}
require an agreement among objects on a common protocol for message interchange,
our approach does not, being capable of perceiving any message on any object
without neither modifying its class source code nor requiring any common message
protocol.

\subsection{Situations}

A situation is an interesting event for an agent. When a situation occurs, an
agent can react quickly or analyze carefully its future actions. Interesting
situations are detected by a \texttt{SituationManager} object when a message
is perceived or a communication is received by taking into account an agent's
mental states such as beliefs, knowledge and goals.

Brainstorm/J represents situations by means of a set of logic clauses contained
into a logic module~\cite{amandi99:multiParadigm}. As an example, we show
how to define a simple situation called \emph{boxInFront}. This situation occurs
when a box is at~\( (X,Y) \) such that~\( (X,Y) \) is in the front of a
robot. This situation is expressed as follows:

\begin{lyxcode}
situation(boxInFront,Box)~:-~

~~location(box(Box),X,Y),~\#Box~is~at~X,~Y

~~newInstance('java.awt.Point',{[}X,Y{]},Front),~\#Front=(X,Y)

~~baseObject(Base),~\#Base~is~the~base~object(Forklift)

~~send(Base,nextLocation,{[}{]},Front).
\end{lyxcode}
in this clause, location(box(\emph{Box}), \emph{X}, \emph{Y}) denotes that \emph{Box}
is at~\( (X,Y) \); newInstance('java. awt.Point', {[}\emph{X}, \emph{Y} {]},
\emph{Front}) means that \emph{Front} is an instance of \texttt{Point} equals
to~\( (X,Y) \); and baseObject(\emph{Base}) means that \emph{Base} is the
base object of an agent. As a consequence, situation(boxInFront, \emph{Box})
is true only if \emph{Box} is at front of a robot.

\subsection{Deliberation}

Brainstorm prescribes an architectural component for dealing with deliberation:
carefully analyzing future actions in order to achieve a set of goals. The implementation
of this component has been done taking into account that agents should be able
to perform several activities concurrently. For example, an hybrid agent should
be able to perceive its environment reacting accordingly while engaged in a
dialog with other agents and reasoning how to achieve its goals. Moreover, agents
should be able to reason about how to achieve their goals by using several concurrent
reasoning mechanisms and taking into account the effects of environmental changes
in the reasoning process and dependencies among different reasoning mechanisms.
For example, while building a plan, a belief can change as a consequence of
perception; an agent should analyze how this new belief affect the plan, repairing
it if necessary and analyzing possible dependencies among its internal tasks.

To support these features, Brainstorm/J assigns to every internal agent component
its own thread. Moreover, the deliberative process is performed by several concurrent
objects that are able to: perceive any activity within an agent such as a commitment
to a goal, a new perception, an achieved goal, a produced plan, etc\footnote{%
There are more than~30 different events related to the internals of an agent
which can be used by reasoning objects.
}; produce actions and partial plans; take an already produced action or plan
to execute or modify it, interact with other objects by means of events such
as kill, wait, achieve a goal, etc\footnote{%
There are more than~20 different events related to interaction among reasoning
objects.
}.

Brainstorm/J defines several types of objects for reasoning. Among them, the
most important are: \texttt{PlanProducerKS} produces a partial plan for achieving
a goal, class \texttt{DelibStrKS} uses a generic planning algorithm as GraphPlan~\cite{blum97:graphplan}
or any user defined algorithm for producing partials plans, \texttt{Executor}
executes actions or plans produced by other objects, \texttt{AbsPlan} groups
related objects into partial and incomplete plans representing a course of action
for achieving a goal, \texttt{PlanAdapter} defines generic services for adapting
partial plans, \texttt{DistanceReductionKS} produces actions for reducing the
distance to a goal, etc.

To clarify the concepts introduced in this section we describe the design of
some classes of Brainstorm/J which define a simple deliberative mechanism. A
deliberative agent decide which actions execute based on its goals. There are
many decision mechanisms that can be used to accomplish this task, for example,
off-line planning, on-line planning or rules. A tool can provide a highly generic
mechanism such as a planning algorithm to do so. Quite certainly in some applications
the algorithm would not be capable of handling application specific constraints
or to take into account information that would lead to a better performance.
The approach used to design Brainstorm/J is different. Instead of providing
a generic mechanism to deliberate, it defines a flexible infrastructure to build
decision procedures for goal-directed behavior.

\begin{figure}
{\par\centering \resizebox*{1\columnwidth}{!}{\includegraphics{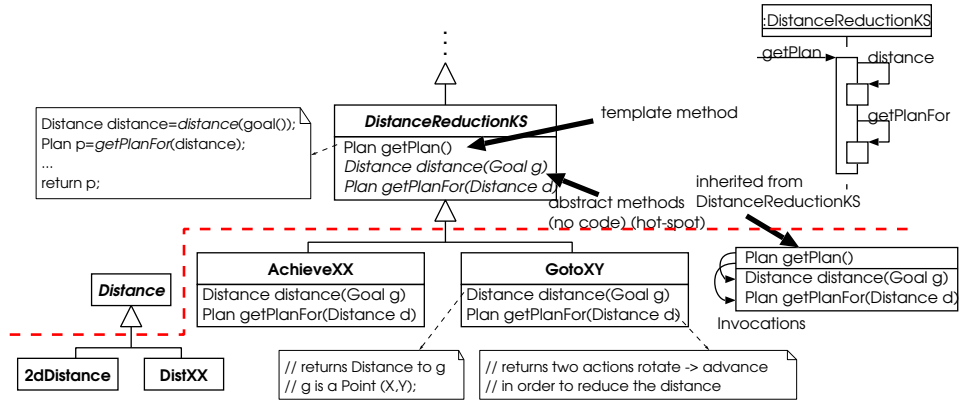}} \par}

\caption{Reusing agent behavior defined in template methods\label{fig:template1}}
\end{figure}

Figure~\ref{fig:template1} shows a diagram with some classes of Brainstorm/J
in charge of deliberation. The abstract class \texttt{DistanceReductionKS} can
be used by a deliberative agent to achieve its goals. The class defines a template
operation named \texttt{getPlan} to reduce the \emph{distance} to a goal\footnote{%
For example, an agent in a two dimensional grid can use the Euclidean distance
between its actual position and its goal, an agent in Internet can measure the
number of sites between its actual position and its target taking into account
the traffic, delays, etc.
}. The class also defines two abstract operations: \texttt{distance} and \texttt{getPlanFor}.
The idea here is that different agents will have completely different ways of
measuring the distance to a goal and to build a plan in order to reduce that
distance, so these are hot-spots. Note that the \texttt{getPlan} operation invokes
these two hot-spots in order to build a plan\footnote{%
This is different from traditional~AI planning, since the output of the algorithm
is not a complete plan to achieve a goal, but \emph{steps} (set of actions)
to reduce the \emph{distance} between the agent and its goal. This differs from
IA planning from that a step is built and executed until the goal is achieved
or dropped, unlike traditional IA planning, execution and planning are interleaved.
}. As a consequence, this method can be adapted by defining the two abstract
methods (hot-spots).

A subclass of \texttt{DistanceReductionKS} that defines the two abstract methods
will be able to use the \texttt{getPlan} method. For example, \texttt{GotoXY}
defines \texttt{distance} to calculate the distance between an agent and a given
point in a two dimensional space; \texttt{getPlanFor} produces two actions (rotate
and advance one step) in order to reduce that distance. Therefore, multiple
invocations of the \texttt{getPlan} operation will produce a complete course
of actions to achieve a goal.

As shown in the example, programmers using Brainstorm/J deal only with application
dependant functionality by defining abstract methods. This is as a consequence
of the usage of template methods defining the common functionality of agents.

\subsection{Reactive behavior}

When an interesting situation is detected by a \texttt{SituationManager} object,
an agent can react in a predefined manner. In order to react, an agent uses
a \texttt{Reactor} object and a set of reactions (class \texttt{Reaction}). 

\begin{figure}
{\par\centering \resizebox*{0.95\columnwidth}{!}{\includegraphics{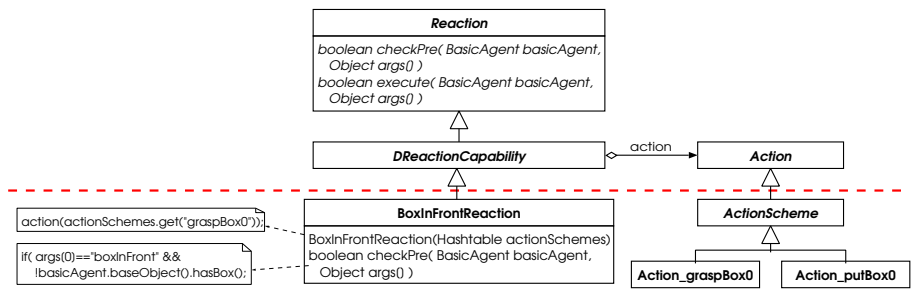}} \par}

\caption{A reaction to the \emph{boxInFront} situation\label{fig:reactor}}
\end{figure}

A \texttt{Reaction} is composed of two parts: a precondition and an agent's
basic skill represented by a method of the base object. When a situation occurs,
the \texttt{Reactor} object checks all the reactions executing only those whose
precondition is true. Figure~\ref{fig:reactor} shows a reaction called \texttt{BoxInFrontReaction}
to the \emph{boxInFront} situation defined previously. This reaction executes
an action named \emph{graspBox0} as a response to \emph{boxInFront} situation.
The action \emph{graspBox0} succeeds if the agent is not holding a box. As a
result of the execution of \emph{graspBox0}, the box is no longer on the floor,
but is held by the agent.

\subsection{Communication and coordination}

Brainstorm/J supports the construction of agents which communicate robustly
over a~LAN by using several protocols such as~e-Mail, TCP/IP or HTTP; diverse
agent communication languages such as IL~\cite{demazeau95:fromInterCollec},
KQML~\cite{finin94:kqml} or FIPA's ACL~\cite{fipa97:spec}; and multi-agent
coordination through COOL-like~\cite{barbuceanu97:coolDesign} conversations.

An agent composed by a \texttt{Communicator} object is able to interact with
other agents. This class only defines an abstract interface for communication,
so programmers should extend it according to their requirements. In addition,
Brainstorm/J provides a concrete class for communication by using KQML. The
concrete class \texttt{KQMLCommunicator} is able to send KQML messages through
the Internet by using TCP/IP, SMTP/POP (e-Mail), FTP or HTTP. 

Services provided by \texttt{KQMLCommunicator} are built by using JATLite~\cite{petrie96:jatlite}.
As a consequence, all communications pass through a centralized component called
Agent Message Router (AMR). The~AMR also provides name services, brokering
and off-line operations.

Brainstorm/J provides several classes for responding to KQML performatives in
a default manner. For example, class \texttt{AskOneHandler} can be used for
responding to all \texttt{ask-one} messages in a given content language and
ontology. This class verifies whether the content of an \texttt{ask-one} message
can be deduced from agent's mental state and responds accordingly.

The framework supports coordination by means of COOL~\cite{barbuceanu97:coolDesign}
conversations. A conversation is specified by a \emph{conversation class} and
a number of \emph{conversation rules}. A conversation class defines a set of
interactions among agents. Each possible interaction in a given point of a conversation
is described by conversation rules.

\begin{figure}
{\par\centering \subfigure[A simple conversation class\label{fig:firstQueenConv}]{\resizebox*{0.43\columnwidth}{!}{\includegraphics{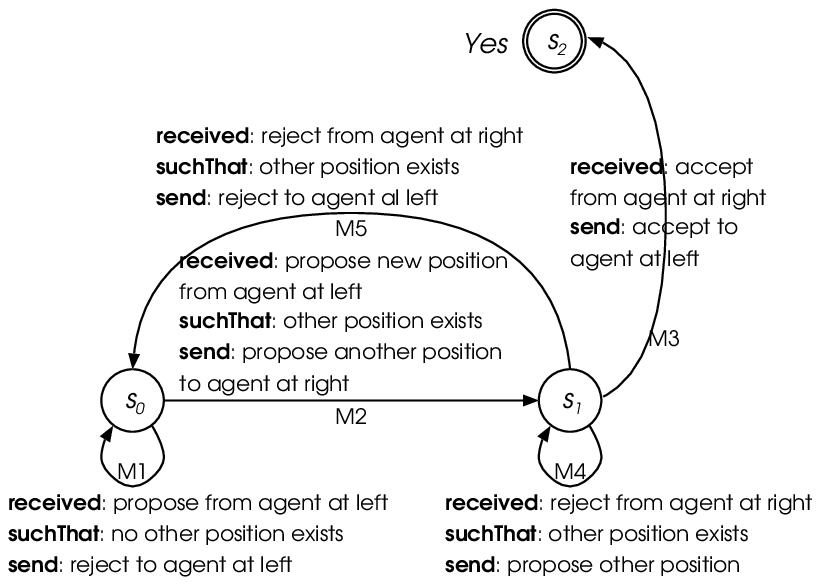}}} ~~~~~\subfigure[Conversation classes and conversation rules\label{fig:conversation}]{\resizebox*{0.5\columnwidth}{!}{\includegraphics{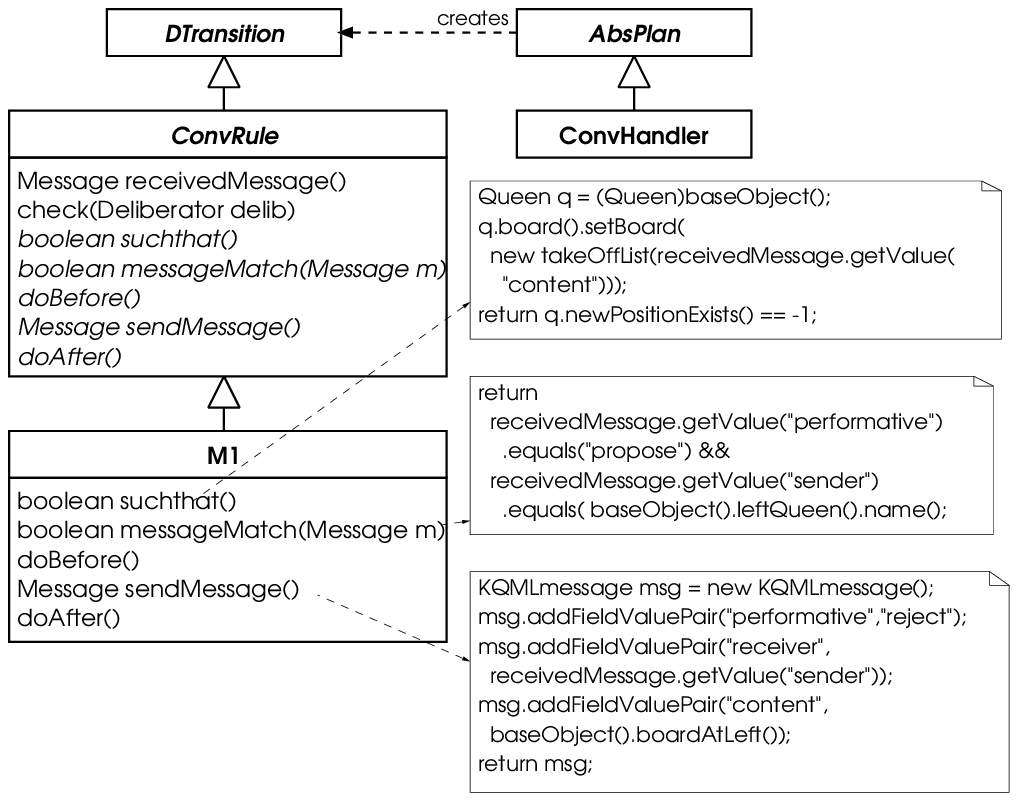}}} \par}

\caption{Coordination using COOL-like conversations}
\end{figure}

A conversation can be represented as a definite finite automata (DFA). Each
state of the~DFA represents a state of a conversation; each transition is equivalent
to a conversation rule. For example, Fig.~\ref{fig:firstQueenConv} shows a
simple conversation class belonging to a multi-agent implementation of the \( n \)-Queens
problem~\cite{chauhan98:jafmas,zunino2000:brainstormJ}.

A conversation rule relates two states of a conversation. As shown in Fig.~\ref{fig:conversation},
a conversation rule occurs when some event activates that transition (\texttt{suchThat}
method). When a transition occurs, actions attached to it are executed (\texttt{doBefore}
and \texttt{doAfter} methods) and a message is sent (\texttt{sendMessage} method).
This process is described by the \texttt{check} template method of the class
\texttt{ConvRule}. Figure~\ref{fig:conversation} shows a class named \texttt{M1}
which defines the abstract methods of \texttt{ConvRule} to represent the rule
M1 of the conversation shown in Fig.~\ref{fig:firstQueenConv}.

\section{Experiences using the Framework\label{sec:Experiences}}

Brainstorm/J has been used to develop two multi-agent systems: Forklifts and
\( n \)-Queens. The first MAS consists of a set of forklifts which tries to
arrange a set of boxes. This MAS has been used to test different agents' behaviors
such as random walker, reactive planning and heuristics for conflict resolution.
Moreover, Forklifts has been used to test several agent architectures such as
Interrap~\cite{Muller96:DIA} and Brainstorm~\cite{amandi98:metaArchitecture}.

On the other hand, the~\( n \)-Queens system is a multi-agent solution to
the \( n \)-Queens problem. In this~MAS, each agent represents a queen with
the sole objective of finding a safe position. Each agent is assigned to a column
and is able to move freely along it. Furthermore, agents communicate among them
by using~KQML.

The~\( n \)-Queens MAS has been used for comparing Brainstorm/J with other
agent framework called JAFMAS~\cite{chauhan98:jafmas}. In order to compare
both implementations we have used several source code metrics such as non commenting
source statements (NCSS), number of methods and classes, NCSS per class, NCSS
per method and cyclomatic complexity number~(CCN)~\cite{mccabe76:complexity}.
Table~\ref{tab:queensMetrics1} summarizes the results obtained by these code
metrics. In the column \emph{difference} we can observe that the implementation
made with Brainstorm/J is simpler and shorter in terms of source code, than
the one made with JAFMAS. It is important to take into account that these results
are not concluding, since they only show a tendency in favor of Brainstorm/J.

Related to performance, both implementations shown very similar numbers, despite
the usage of meta-objects in Brainstorm/J. This is a consequence of the usage
of the framework JMOP, which introduces invocations to the meta-level by modifying
Java byte-codes, and meta-object managers~\cite{campo99:luthier} to manage
the associations among objects and meta-objects.

\begin{table}

\caption{Comparison between two implementations of the \protect\( n\protect \)-Queens
problem\label{tab:queensMetrics1}}

{\small
\vspace{0.3cm}
{\centering \begin{tabular}{c|D..{2.3}|D..{2.3}|D..{2.3}}
\hline 
&
\multicolumn{1}{|c|}{\textbf{JAFMAS}}&
\multicolumn{1}{|c|}{\textbf{Brainstorm/J}}&
\multicolumn{1}{|c}{\textbf{Difference}}\\
\hline 
Classes&
18&
21&
14.28\%\\
Methods&
92&
87&
-5.43\%\\
NCSS&
625&
491&
-21.44\%\\
Methods per class&
5.11&
4.14&
-18.98\%\\
NCSS per class&
34.72&
23.38&
-32.66\%\\
NCSS per method&
6.79&
5.64&
-16.93\%\\
Average CC per method&
1.57&
1.51&
-3.82\%\\
\hline 
\end{tabular}\par}\vspace{0.3cm}

}
\end{table}

\section{Related Work\label{sec:Related}}

There are currently several tools for building agents. Among them, we can mention
AgentBuilder~\cite{reticular99:AgentBuilder}, DECAF~\cite{graham99:decaf},
JAF~\cite{horling98:jaf}, JAFMAS~\cite{chauhan98:jafmas} and JAFIMA~\cite{kendall99:framework}.

AgentBuilder~\cite{reticular99:AgentBuilder} and DECAF~\cite{graham99:decaf}
are toolkits for building MAS based on a set of Java classes which implement
common services such as communication, coordination and reasoning. Although
they provide some mechanisms for adding user-defined code, they lack the flexibility
and adaptability provided by our framework. JAFMAS~\cite{chauhan98:jafmas}
is a Java framework for building MAS based on a COOL-like model of coordination.
As the other tools it cannot be extended.

JAFIMA (Java Framework for Intelligent and Mobile Agents)~\cite{kendall99:framework}
takes a different approach from the other tools: it is primarily targeted at
expert developers who want to develop agents from scratch based on the abstract
classes provided, so the programming effort is greater than in the other tools.
The weakest point of JAFIMA is its rule-based mechanism for defining agents'
behavior. This mechanism does not support complex behaviors such as on-line
planning or learning. Moreover, the abstractions for representing mental states
lack flexibility and services for manipulating symbolic data.

\begin{figure}
{\par\centering \resizebox*{0.6\columnwidth}{!}{\includegraphics{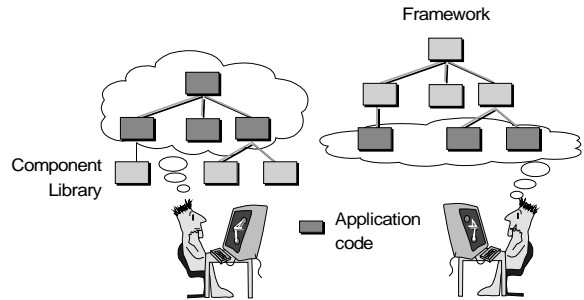}} \par}

\caption{Differences between component libraries and frameworks\label{fig:frame-lib}}
\end{figure}

At this point, we would like to emphasize the differences between frameworks
and another related technique for software reuse: component libraries. Component
libraries are being used to support agent development in tools like JAF~\cite{horling98:jaf}.
A component library consist of a number of reusable program building blocks
such as C~functions, classes or JavaBeans. As shown in Fig.~\ref{fig:frame-lib},
developers using a component library should combine a number of components in
order to build an application. This is achieved by writing algorithms\footnote{%
The composition can be assisted by a visual tool.
} that call methods or functions provided by the component library in order to
establish interactions and collaborations among these components.

A framework as Brainstorm/J is not just a collection of components but also
defines a generic design. When programmers use a framework they reuse that design
and save time and effort. In addition, because of the bidirectional flow of
control frameworks can contain much more functionality than a traditional library
regardless if it is a procedural or class library~\cite{fayad97:frameCACM}.

None of the analyzed tools in this section handle concurrency or asynchronism
within agents. On the other hand, agents built with Brainstorm/J are able to
maintain several conversations, plan how to achieve its goals and perceive its
environment, all of them concurrently and taking into account possible interdependencies
and conflicts among these process.

\section{Conclusions and Future Work\label{sec:Conclusions}}

In this paper a framework for MAS named Brainstorm/J has been described. This
framework is based on the Brainstorm architecture. As a result, it supports
different types of agents with capabilities such as perception, communication,
symbolic manipulation of mental states, mobility, reaction and deliberation.

The framework has been built by using a support for meta-objects for Java named
JMOP and a multi-paradigm language named JavaLog. It is worth noting that meta-objects
provide a convenient and flexible mechanism to combine agents' capabilities
to develop different types of agents. Moreover, it is possible to integrate
new agents' capabilities into the framework by adding meta-object classes. 

Brainstorm/J defines the common functionality of agents in abstract classes.
As a result, programmers deal only with application specific functionality,
since the functionality common to all types of agents is reused from the framework.
In this way, the design/programming effort involved in the construction of multi-agent
systems is greatly reduced. 

Unlike other tools, Brainstorm/J is able to support many types of agents in
diverse application domains, since it is possible to extend and adapt the framework
by means of subclassification.

Future research on Brainstorm/J will integrate single-agent learning and multi-agent
learning into the framework by materializing the learner component prescribed
by Brainstorm. In addition a methodology for designing and programming agents
with Brainstorm/J should be developed.

\bibliographystyle{apalike}
\bibliography{yves,learning,reflection,java,spec,reactive,agentTools,bdi,amandi,planning,logic,frameworks,agents,mobile,arch,scheduling}

\end{document}